%% file: ISVC2018_DVE.tex
\documentclass[runningheads]{llncs}
\usepackage{graphicx}
\usepackage{amsfonts}
\usepackage{epsfig} 
\usepackage{amsmath} 
\usepackage{amssymb}  
\usepackage{float}
\usepackage{array}
\usepackage{color}
\usepackage{citesort}
\usepackage{flushend}
\usepackage{url}
\usepackage{algorithm}
\usepackage[compatible]{algpseudocode}

\algnewcommand\AAND{\textbf{ and }}
\algnewcommand\Or{\textbf{ or }}

\DeclareMathAlphabet{\pazocal}{OMS}{zplm}{m}{n}

\newcommand{\Ws}{\pazocal{W}}
\newcommand{\Ns}{\pazocal{N}}

\newcommand{\Bs}{\pazocal{B}}
\newcommand{\Cs}{\pazocal{C}}
\newcommand{\Ds}{\pazocal{D}}

\newcommand{\Ls}{\pazocal{L}}
\newcommand{\Vs}{\pazocal{V}}

\newcommand*{\vertbar}{\rule[-1ex]{0.5pt}{3.25ex}}

\newcolumntype{C}[1]{>{\centering\arraybackslash}p{#1}}
\newcolumntype{M}[1]{>{\raggedright\arraybackslash}p{#1}}

\usepackage{array} 
\newcolumntype{L}[1]{>{\raggedright\let\newline\\\arraybackslash\hspace{0pt}}m{#1}}	
\newcolumntype{S}[1]{>{\centering\let\newline\\\arraybackslash\hspace{0pt}}m{#1}}
\newcolumntype{R}[1]{>{\raggedleft\let\newline\\\arraybackslash\hspace{0pt}}m{#1}}

\makeatletter
\renewcommand*{\@opargbegintheorem}[3]{\trivlist
  \item[\hskip \labelsep{\bfseries #1\ #2}] \textbf{(#3)}\ }
\makeatother

\begin{document}
\title{Vision--Depth Landmarks and Inertial Fusion for Navigation in Degraded Visual Environments}
\titlerunning{Vision--Depth Landmarks and Inertial Fusion for Navigation}
%
\author{Shehryar Khattak \and Christos Papachristos \and Kostas Alexis}
%
%
\institute{Autonomous Robots Lab, University of Nevada, Reno, NV, USA
\url{http://www.autonomousrobotslab.com}}
\maketitle              
\begin{abstract}
This paper proposes a method for tight fusion of visual, depth and inertial data in order to extend robotic capabilities for navigation in GPS--denied, poorly illuminated, and textureless environments. Visual and depth information are fused at the feature detection and descriptor extraction levels to augment one sensing modality with the other. These multimodal features are then further integrated with inertial sensor cues using an extended Kalman filter to estimate the robot pose, sensor bias terms, and landmark positions simultaneously as part of the filter state. As demonstrated through a set of hand-held and Micro Aerial Vehicle experiments, the proposed algorithm is shown to perform reliably in challenging visually--degraded environments using RGB-D information from a lightweight and low--cost sensor and data from an IMU.
\end{abstract}
%
%
%

\section{INTRODUCTION}\label{sec:intro}
\input{2Introduction.tex}

\section{RELATED WORK}\label{sec:related}
\input{3RelatedWork.tex}

\section{PROPOSED APPROACH}\label{sec:approach}
\input{4ProposedMethod.tex}
\section{EXPERIMENTAL EVALUATION}\label{sec:experiments}
\input{5ExperimentalEvaluation.tex}

\section{CONCLUSIONS}\label{sec:concl}
A method for common visual and depth data features alongside their fusion with IMU cues in order to enable autonomous localization in degraded visual environments and specifically low--light, dark, and textureless conditions was proposed. The focus is on an approach that exploits lightweight and ubiquitous RGB-D sensors and therefore can be integrated onboard small systems such as Micro Aerial Vehicles. A set of experimental evaluation studies are presented and demonstrate the ability of the system to provide reliable localization and mapping data in sensing--degraded conditions of darkness and low--light flight.

\bibliographystyle{ieeetr}
\bibliography{ISVC2018_DVE}
\end{document}

%% file: 2Introduction.tex
Robotic systems are being integrated in an increasingly large variety of applications such as infrastructure inspection~\cite{NIR_ICUAS_2017,RHEM_ICRA_2017,mascarich2018multi,papachristos2019autonomous,VSEP_ICRA_2018}, monitoring and surveillance~\cite{grocholsky2006cooperative}, search and rescue~\cite{balta2017integrated} and more. However, in many critical tasks robots have to be able to cope with difficult conditions that challenge their autonomy. Of particular interest are cases involving GPS--denied Degraded Visual Environments (DVEs) that stress the abilities of onboard perception especially for Simultaneous Localization And Mapping (SLAM). Examples of relevant applications include navigating, mapping and characterization of underground settings, as well as search and rescue missions.
\par\noindent
The focus of this work is on enabling reliable robotic autonomy in poorly illuminated, textureless and GPS--denied environments through the use of lightweight and ubiquitous sensing technology; specifically, miniaturized RGB-D sensors and Inertial Measurement Units (IMU). For this purpose we first develop a methodology for fusion of visual and depth information at the feature detection and descriptor extraction levels followed by the integration of these multimodal features with inertial sensor information in an Extended Kalman Filter (EKF) fashion. This approach reflects the fact that small RGB-D sensors have limited range, and is different from a) methods that fuse LiDAR and visual camera data in a loosely--coupled manner, b) RGB-D SLAM algorithms that depend on vision as prime sensing modality, or c) technological approaches that integrate onboard illumination or night vision.
\par\noindent
\begin{figure}[h!]
\centering
  \includegraphics[width=0.99\columnwidth]{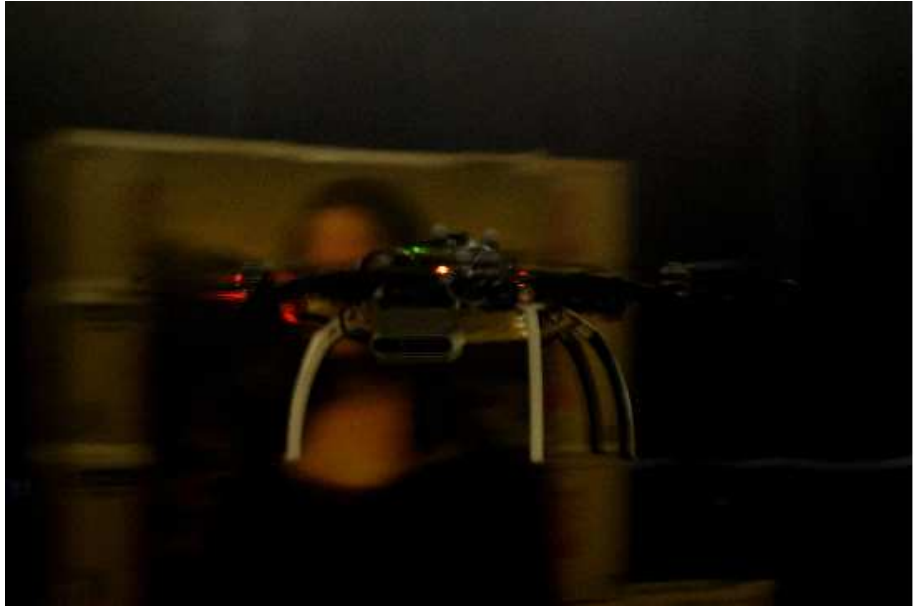}
\caption{Aerial robotic operation in low--light GPS--denied conditions. An Intel Realsense D435 Depth Camera is integrated onboard a Micro Aerial Vehicle of the F450 class.}
\label{fig:mainfigure}
\end{figure}
To verify the proposed solution, a set of experimental studies were conducted including a) a proof--of--concept hand-held validation test inside a dark room where visual information is extremely limited, and b) an aerial robot trajectory in very low--light conditions. Comparison with ground truth and reconstructed maps are presented to demonstrate the reliability of the proposed localization system and its performance in such visually--degraded conditions.
\par\noindent
The remainder of the paper is structured as follows: Section~\ref{sec:related} provides an overview of related work. The proposed approach for common visual and depth features alongside fusion with IMU is presented in Section~\ref{sec:approach}, followed by experimental evaluation in Section~\ref{sec:experiments}. Finally, conclusions are drawn in Section~\ref{sec:concl}.

%% file: 3RelatedWork.tex
Methods for odometry estimation and SLAM have seen rapid growth in the recent years and employ a multitude of sensing modalities. For camera-based systems a set of feature--based~\cite{ORBSlam,libVISO,VOtutorialPart1}, and semi--direct~\cite{SVO} visual odometry algorithms have been proposed. Furthermore, visual--inertial fusion approaches have presented increased robustness and reliability especially when dynamic motion is considered~\cite{ROVIO,leutenegger2015keyframe}.
Although visual odometry techniques have seen great growth in variety and robustness, yet the fact remains that all these techniques rely on proper scene illumination and availability of texture for their operation. 
On the other hand, direct depth sensors such as Light Detection And Ranging (LiDAR) and dense depth sensor are robust against illumination changes or lack of texture.
Light Detection And Ranging (LiDAR) units can produce depth measurements at long ranges and return data in the form of sparse point clouds.
Techniques using LiDAR along with IMU integration to generate odometry estimates have shown very robust results over long ranges~\cite{LOAM} but tend to suffer when matched with short--range sensors and/or when they operate in structure-less environments where geometric constraints are not enough to constrain the underlying optimization process~\cite{zhang2016degeneracy}.
Similarly, dense depth sensors produce dense depth data at short ranges and can be easily combined with RGB images. Techniques such as~\cite{tumRGBDodom,rtabMap,tumRGBDmap}, have shown good odometry estimation results when using RGB-D data.
However, despite the fact that these methods take advantage of the availability of direct depth estimates, on closer inspection it can be noted that handling of depth and visual data is done separately. In~\cite{tumRGBDodom,rtabMap} feature detection and matching is done solely on the visual image for odometry estimation and depth data is utilized for scale estimation and mapping purposes.
Similarly, visual approaches that deal with cases of poor illumination usually do not integrate any other type of information and require an external light source~\cite{alismail,aerconf2018}.
Due to this separate handling of the two sensing modalities, the overall odometry estimation remains prone to illumination changes and lack of texture. To remedy these problems, recent work~\cite{BRAND,RISAS} proposes to encode visual and depth information on the feature detection and descriptor level. Although these approaches improve the robustness in large illumination changes, they are sensitive to the quality of depth data and can become computationally burdensome for real time operations.
\\
Motivated by the discussion above, in this work we present an EKF framework that fuses inertial, visual and depth information for odometry estimation. We use a robot-centric formulation and use inertial measurements to predict feature pixel positions between frames and use the re-projection error as an innovation term for the update step. Visual and depth information are encoded at the feature detector and descriptor level making them more robust in certain DVEs, i.e. low illumination and texture-less conditions. Only a small number of features are tracked as part of the filter state making the whole odometry computationally tractable for real time on--board robot navigation tasks. To the best of our knowledge this tightly integrated multimodal framework has no precedent in the robot odometry estimation literature.

%% file: 4ProposedMethod.tex
\begin{figure}[h!]
\centering
  \includegraphics[width=0.99\columnwidth]{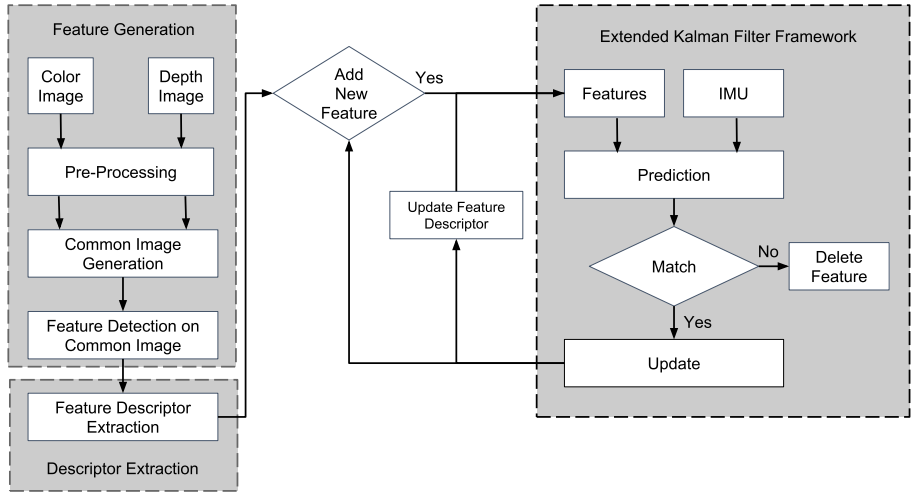}
\caption{An overview of the proposed approach.}
\label{fig:overview}
\end{figure}

Our proposed approach consists of three main components, namely a) Visual--Depth Feature generation, b) Descriptor Extraction and c) inertial fusion using an EKF, as shown in Figure~\ref{fig:overview}. 
%
%
\subsection{Feature Detection}
\begin{figure}[h!]
\centering
  \includegraphics[width=0.99\columnwidth]{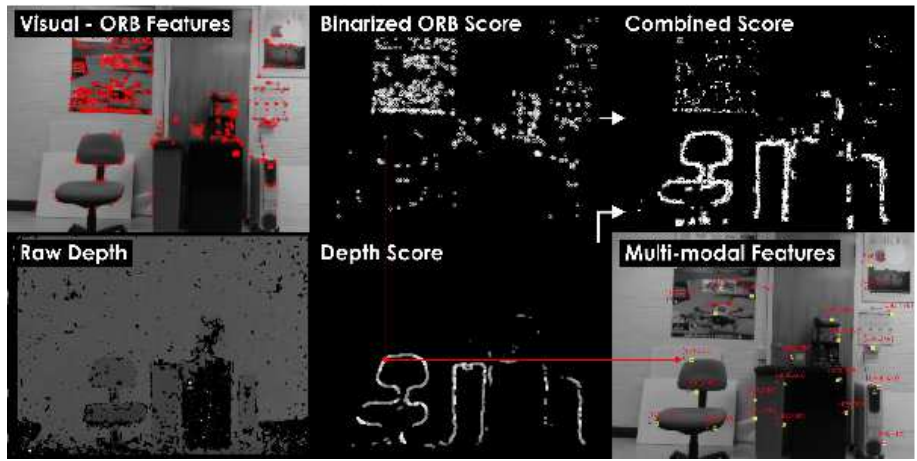}
\caption{Indicative example with steps of the process for common visual and depth feature detection. With red line, the case of one of the multimodal features is presented.}
\label{fig:commonfeatures}
\end{figure}

Upon receiving a pair of visual and depth images our method generates a combined score image that identifies keypoints in both sensing modalities by making use of the intuition that edges in both domains often lie along same image coordinates~\cite{levinson2013}. For this purpose we perform registration between the two images using the intrinsic and extrinsic calibration parameters of the cameras. 
For detection of features in the visual image we make use of the ORB feature detector~\cite{ORB} due to its robustness to image noise and scale invariant properties. We threshold detected ORB points by imposing a quality metric $\lambda$, representing the minimum acceptable Harris corner score, to ensure repeatable detection in low light conditions. We normalize the scores of the remaining visual keypoints and annotate them to create a visual score map $\Vs_{s}(x)$ which represents the score of a visual keypoint at pixel location $x$. For identification of key points in the depth image we calculate the second derivative at each pixel location by applying the Laplacian operator to the image. This allows us to identify edges in the depth domain, as well as to differentiate between points that lie near object edges as part of background or foreground since points in the background take a negative value. Next, we filter out noisy edge points by comparing the score at the pixel location to the depth error of the neighbouring pixels used in the calculation, which is a quadratic function of depth as mentioned in~\cite{realsense}. This operation is defined as:
\small
\begin{eqnarray}
 \Ds_{s}(x) = \max\left(\Ls(x) - \sum_{\textrm{i=$1$}}^{n}\Ns(i),0\right)
\end{eqnarray}
\normalsize
where $\Ds _s(x)$ is the score at depth pixel location $x$, $\Ls(x)$ is the result of applying Laplacian operation at the pixel location and $\Ns(i)$ is the depth error at pixel location $i$ among the $n$ neighboring pixels used in the calculation. Furthermore, to reduce the number of points along the edges and to identify corner points, we apply non-maxima suppression and only keep points with a gradient direction in the range of $30^\circ\leq\theta\leq60^\circ$ in each quadrant. We normalize the scores to get our final depth score map $\Ds_{s}(x)$. The visual and depth score maps are then combined into a single score map which allows the method to identify multimodal keypoints and also maintain the best keypoints from each modality. This is given as:
\small
\begin{eqnarray}
 \Cs_{s}(x) = \min(
\gamma\Vs_{s}(x) + \left(1-\gamma\right)\Ds_{s}(x),{s_\textrm{sat}})
\end{eqnarray}
\normalsize
where $\Cs_{s}(x)$ represents the combined score at every pixel location, $\gamma$ defines the contribution factor of each sensing modality, and $s_\textrm{sat}$ is a fixed value used to saturate the value of $\Cs_{s}(x)$. The best keypoints from the combined score map are selected using an adaptive Euclidean distance constraint to ensure the method balances having enough features tracked in the filter while maintaining a distribution of features across the image frame. Figure~\ref{fig:commonfeatures} illustrates an example of this process for common feature detection. 
%
%

\subsection{Descriptor Extraction}
Given a set of multimodal keypoints we extract a descriptor that utilizes both visual and depth information. For this purpose we chose to use the binary descriptor BRAND~\cite{BRAND} because of its ability to encode visual and depth information, while maintaining fast performance during the descriptor extraction and matching processes. 
BRAND generates two binary strings that encode the visual and depth neighborhood information of keypoints individually, by performing pair--wise intensity and geometric tests on visual and depth images respectively. These bit strings are then combined using a bit--wise $OR$ operation to create a combined descriptor. To generate the visual part of the descriptor, for a pair of pixel locations $P_{1}$ and $P_{2}$ with intensities $I_{1}$ and $I_{2}$, BRAND performs the following visual comparison $V$:
\small
\begin{eqnarray}\label{eqs:visual_eq}
V \left(P_{1},P_{2}\right)=
\begin{cases} 
1, \quad\textrm{if}\ I_{1}<I_{2}
\\
0, \quad\textrm{otherwise}
\end{cases} 
\end{eqnarray}
\normalsize
However, differing from the original work, our method modifies the above comparison for two reasons. First, comparing pixel intensities directly is susceptible to pixel noise and can cause erroneous bit switching in the descriptor. Secondly, in poorly illuminated conditions, visual camera sensors are susceptible to dark noise which can generate false intensity values that
can introduce noise in the visual part of the descriptor. To reduce the sensitivity of the descriptor to pixel noise, instead of comparing pixel intensity values we compare mean intensity values using patches of size $9\times9$ created around sampled pair locations. Secondly, to reduce the effect of dark noise we subtract a small intensity value, representing dark noise, from the mean intensity values of the patches before performing the pair-wise comparison. This intensity representation of dark noise $\left(I_{\textrm{DN}}\right)$ can be calculated by collecting images in a very dark environment where we expect the intensity value to be zero and deriving the mean intensity value across these images. The modified intensity comparison function for a pair of pixel locations $P_{1}$ and $P_{2}$ with mean patch intensities $\overline{I}_{1}$ and $\overline{I}_{2}$ takes the form:

\scriptsize
\begin{eqnarray}\label{eqs:visual_eq_mod}
V \left(P_{1},P_{2}\right)=
\begin{cases} 
1, \quad\textrm{if}\ \ \max\left(0,\overline{I}_{1}-I_{\textrm{DN}}\right)<\max\left(0,\overline{I}_{2}-I_{\textrm{DN}}\right) \\ 
0, \quad\textrm{otherwise}
\end{cases}
\end{eqnarray}
\normalsize
The $\max$ function in the above equation ensures that the minimum allowed intensity value is $0$. The depth part of the descriptor is maintained as describe in~\cite{BRAND}.

\subsection{Extended Kalman Filter for IMU Fusion}
We fuse inertial information with our multimodal features (landmarks) by tracking them as part of the state of an EKF, where state propagation is done by using proper acceleration $\hat{\mathbf{f}}$ and rotational rate measurements $\hat{\boldsymbol{\omega}}$ provided by an IMU. This formulation allows us to predict the feature locations between successive frames hence reducing the search space for feature matching without the need for feature mismatch pruning. As features are part of the filter state we have an estimate of their uncertainty which we utilize to dynamically scale the search patch for every feature individually for feature matching purposes. Our filter structure is similar to the one proposed in~\cite{ROVIO}. In our formulation, three coordinate frames namely, IMU fixed coordinate frame ($\Bs$), the camera fixed frame $\Vs$, and the world inertial frame $\Ws$, are used. As we register our depth image with respect to the visual image, all the depth data is expressed in the camera fixed frame $\Vs$.
Our multimodal features are expressed in $\Vs$ and are parameterized using a landmark approach which models $3\textrm{D}$ feature locations by using a $2\textrm{D}$ bearing vector $\boldsymbol{\mu}$, parametrized with azimuth and elevation angles, and a depth parameter $\boldsymbol{d}$. By using this parameterization, feature locations in the camera frame and their depth estimates are decoupled which allows us to use multimodal, vision--only and depth--only features interchangeably as part of the state vector.
Hence, the employed state vector with dimension $l$ and associated covariance matrix $\mathbf{\Sigma}_l$ is:
\small
\begin{eqnarray}\label{eq:roviostate}
 \mathbf{x} = [ \underbrace{\overbrace{\mathbf{r}~\mathbf{q}}^\text{pose, $l_p$}\boldsymbol{\upsilon}~\mathbf{b}_f~\mathbf{b}_\omega
 }_\text{robot states, $l_s$}~\vertbar~\underbrace{\boldsymbol{\mu}_0,~\cdots~\boldsymbol{\mu}_J~\rho_0~\cdots~\rho_J}_\text{multimodal features states, $l_f$}]^T
\end{eqnarray}
\normalsize
where $l_p,l_s,l_f$ are dimensions, $\mathbf{r}$ and $\boldsymbol{\upsilon}$ are the robot-centric position and velocity of the IMU respectively, expressed in $\Bs$, $\mathbf{q}$ is the IMU attitude represented as a map from $\Bs \rightarrow \Ws$, $\mathbf{b}_f$  and $\mathbf{b}_\omega$represents the additive accelerometer and gyroscope biases respectively expressed in $\Bs$,
while $\boldsymbol{\mu} _j$ is the bearing vector to feature $j$ expressed in $\Vs$ and $\rho_j$ is the depth parameter of the $j^{th}$ feature such that the feature distance $d_j$ is $d(\rho_j) = 1/\rho_j$. 
Given the estimation of the robot pose, this is then expressed on the world frame $\Ws$ and the relevant pose transformations are available. This enables state feedback control and allows autonomy in difficult DVE cases of darkness and broadly poor illumination and lack of texture.

%% file: 5ExperimentalEvaluation.tex
To evaluate the proposed solution for multimodal sensor fusion
, a visual-depth-inertial perception unit consisting of a lightweight and low--cost Intel Realsense D435 Depth Camera, and a software synchronized VN--100 IMU from VectorNav was employed. 
Intel Realsense D435 provides RGB images as well as reliable depth information in the range from $0.75m$ to $6.0m$.
Beyond intrinsics calibration, camera--to--IMU extrinsics are identified based on the work in~\cite{kalibr}. A set of experimental studies were conducted, in particular a) a hand-held evaluation study inside a dark room, alongside b) an experiment using a small aerial robot operating in low--light conditions. For both studies, the method processes Realsense D435 data at $10\textrm{Hz}$, while IMU updates are received at $200\textrm{Hz}$.

\subsection{Hand-held Evaluation}
\begin{figure}[h!]
\centering
  \includegraphics[width=0.99\columnwidth]{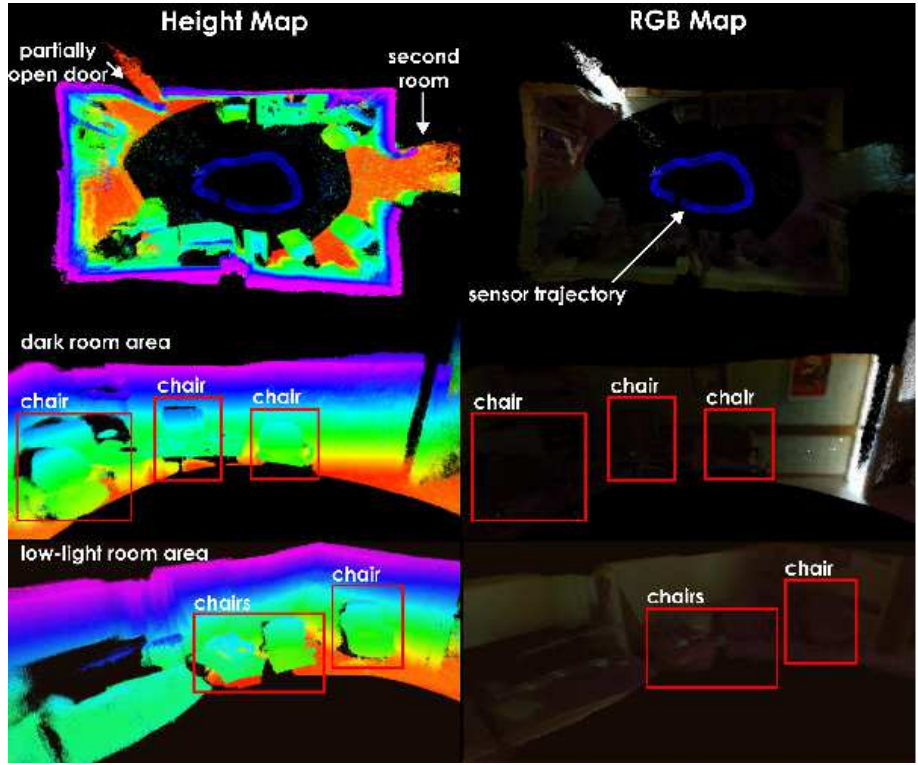}
\caption{Hand-held results regarding the localization inside a dark room. The mapping result indicates the consistency of the estimated trajectory which remains robust even in the most dark and low--light subsets of the environment.}
\label{fig:handheld}
\end{figure}

The first experimental evaluation refers to the localization and mapping inside a $7.6\times 5\times 2.3\textrm{m}$ dark room. Utilizing the Intel Realsense D435 Depth Camera and the VN--100 IMU, the method was found to be able to reliably estimate the motion trajectory and therefore allow consistent reconstruction of the $3\textrm{D}$ map. Most notably, the method maintains robustness even in very dark subsets of the environment, where visual camera data is non--informative and therefore traditional visual or visual--inertial odometry pipelines cannot cope with in a reliable manner. In these areas, furniture provides depth information that allows the framework to work. Figure~\ref{fig:handheld} presents results from this study. 
%

\subsection{Micro Aerial Vehicle Flight Test}
For the flight experiment, a custom--built hexarotor Micro Aerial Vehicle (MAV) is employed and has a weight of $2.6\textrm{kg}$. The system relies on a Pixhawk--autopilot for its attitude control, while integrating an Intel NUC5i7RYH and executing a complete set of high--level tasks on-board with the support of the Robot Operating System (ROS). For the purposes of position control, a Linear Model Predictive Control strategy has been deployed following previous work in~\cite{mpc_rosbookchapter}. 
The robot integrates the same perception unit, i.e., the Intel Realsense D435 Depth Camera and the VN--100 IMU. An instance of the robot during the flight experiment is shown in Figure~\ref{fig:mainfigure}.
\par
\begin{figure}[h!]
\centering
  \includegraphics[width=0.99\columnwidth]{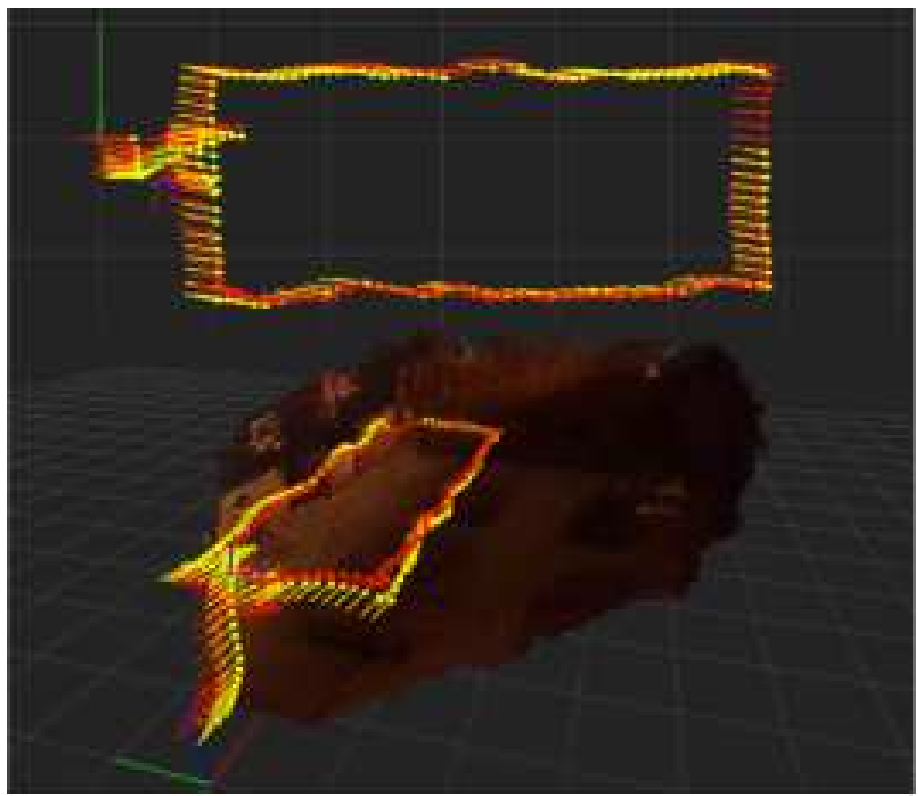}
\caption{Upper plot: The estimated robot trajectory (red) versus ground--truth information using a VICON motion capture system (yellow). As shown, the proposed method provides consistent localization results in dark visually--degraded conditions based on the fusion of visual-depth and inertial data. Bottom plot: reconstructed map of the environment based on the trajectory conducted by the robot. As shown, a major part of the environment presents very low--light conditions. In all plots, it can be identified that the motion capture system presented partial loss of data for segments of the conducted trajectory. A video of this experiment is uploaded at}
\label{fig:flighttest}
\end{figure}

\begin{figure}[h!]
\centering
  \includegraphics[width=0.9\columnwidth]{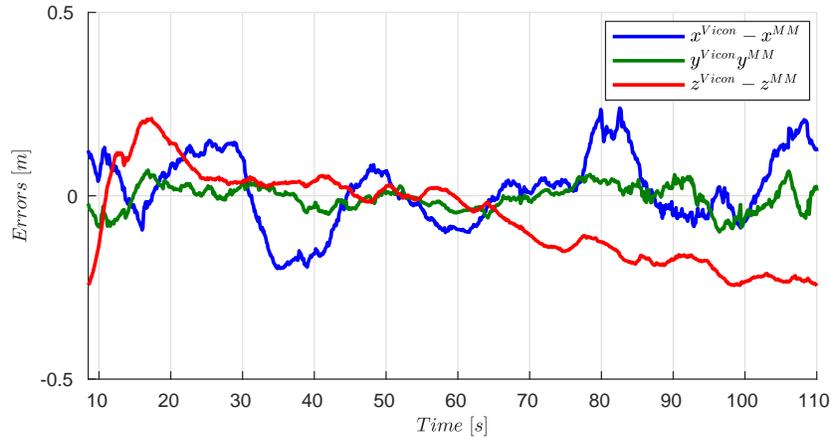}
\caption{Error plot between the onboard estimated trajectory $(\cdot)^{MM}$ and ground--truth provided by a VICON motion capture system. As shown, very small error is achieved despite the operation in dark visually--degraded conditions.}
\label{fig:timeplot}
\end{figure}

\noindent
The conducted experimental study relates to that of tracking a prescribed rectangle trajectory ($\textrm{length}=4.8\textrm{m}$, $\textrm{width}=1.95\textrm{m}$) in a low--light environment. The real--time estimated trajectory is compared against ground--truth information provided by a VICON Motion Capture system. Figure~\ref{fig:flighttest} presents the derived results with the proposed method running on--board the MAV, alongside the reconstructed map of the environment. As shown, the derived trajectory is mostly on par with the ground--truth data and therefore reliable mapping is also achieved. Figure~\ref{fig:timeplot} presents an error plot for each axis of the robot trajectory. The video of the experiment can be seen at (\url{https://tinyurl.com/DVEResults})
%

%

%% file: ISVC2018_DVE.bbl
\begin{thebibliography}{10}

\bibitem{NIR_ICUAS_2017}
{C. Papachristos, S. Khattak and K. Alexis}, ``Autonomous exploration of
  visually-degraded environments using aerial robots,'' in {\em 2017
  International Conference on Unmanned Aircraft Systems}, IEEE, 2017.

\bibitem{RHEM_ICRA_2017}
{C. Papachristos, S. Khattak, and K. Alexis}, ``Uncertainty--aware receding
  horizon exploration and mapping using aerial robots,'' in {\em IEEE
  International Conference on Robotics and Automation}, May 2017.

\bibitem{mascarich2018multi}
F.~Mascarich, S.~Khattak, C.~Papachristos, and K.~Alexis, ``A multi-modal
  mapping unit for autonomous exploration and mapping of underground tunnels,''
  in {\em 2018 IEEE Aerospace Conference}, pp.~1--7, IEEE, 2018.

\bibitem{papachristos2019autonomous}
C.~Papachristos, M.~Kamel, M.~Popovi{\'c}, S.~Khattak, A.~Bircher,
  H.~Oleynikova, T.~Dang, F.~Mascarich, K.~Alexis, and R.~Siegwart,
  ``Autonomous exploration and inspection path planning for aerial robots using
  the robot operating system,'' in {\em Robot Operating System (ROS)},
  pp.~67--111, Springer, 2019.

\bibitem{VSEP_ICRA_2018}
{T. Dang, C. Papachristos, and K. Alexis}, ``Visual saliency-aware receding
  horizon autonomous exploration with application to aerial robotics,'' in {\em
  IEEE International Conference on Robotics and Automation (ICRA)}, May 2018.

\bibitem{grocholsky2006cooperative}
B.~Grocholsky, J.~Keller, V.~Kumar, and G.~Pappas, ``Cooperative air and ground
  surveillance,'' {\em IEEE Robotics \& Automation Magazine}, 2006.

\bibitem{balta2017integrated}
H.~Balta {\em et~al.}, ``Integrated data management for a fleet of
  search-and-rescue robots,'' {\em Journal of Field Robotics}, vol.~34, 2017.

\bibitem{ORBSlam}
R.~Mur-Artal, J.~M.~M. Montiel, and J.~D. Tardos, ``Orb-slam: a versatile and
  accurate monocular slam system,'' {\em IEEE Transactions on Robotics}, 2015.

\bibitem{libVISO}
B.~Kitt {\em et~al.}, ``Visual odometry based on stereo image sequences with
  ransac-based outlier rejection scheme,'' in {\em Intelligent Vehicles
  Symposium (IV)}, 2010.

\bibitem{VOtutorialPart1}
D.~Scaramuzza and F.~Fraundorfer, ``Visual odometry: Part i: The first 30 years
  and fundamentals,'' {\em IEEE robotics \& automation magazine}, vol.~18,
  2011.

\bibitem{SVO}
C.~Forster, M.~Pizzoli, and D.~Scaramuzza, ``Svo: Fast semi-direct monocular
  visual odometry,'' in {\em International Conference on Robotics and
  Automation}, 2014.

\bibitem{ROVIO}
M.~Bloesch, S.~Omari, M.~Hutter, and R.~Siegwart, ``Robust visual inertial
  odometry using a direct ekf-based approach,'' in {\em Intelligent Robots and
  Systems (IROS)}, pp.~298--304, IEEE, 2015.

\bibitem{leutenegger2015keyframe}
S.~Leutenegger, S.~Lynen, M.~Bosse, R.~Siegwart, and P.~Furgale,
  ``Keyframe-based visual--inertial odometry using nonlinear optimization,''
  {\em The International Journal of Robotics Research}, vol.~34, 2015.

\bibitem{LOAM}
J.~Zhang and S.~Singh, ``Loam: Lidar odometry and mapping in real-time,'' in
  {\em Robotics: Science and Systems Conference}, (Pittsburgh, PA), July 2014.

\bibitem{zhang2016degeneracy}
J.~Zhang {\em et~al.}, ``On degeneracy of optimization-based state estimation
  problems,'' in {\em IEEE International Conference on Robotics and
  Automation}, 2016.

\bibitem{tumRGBDodom}
C.~Kerl, J.~Sturm, and D.~Cremers, ``Robust odometry estimation for rgb-d
  cameras,'' in {\em International Conference on Robotics and Automation},
  2013.

\bibitem{rtabMap}
M.~Labbe and F.~Michaud, ``Appearance-based loop closure detection for online
  large-scale and long-term operation,'' {\em IEEE Transactions on Robotics},
  2013.

\bibitem{tumRGBDmap}
F.~Endres, J.~Hess, J.~Sturm, D.~Cremers, and W.~Burgard, ``3-d mapping with an
  rgb-d camera,'' {\em IEEE Transactions on Robotics}, vol.~30, no.~1,
  pp.~177--187, 2014.

\bibitem{alismail}
H.~Alismail, M.~Kaess, B.~Browning, and S.~Lucey, ``Direct visual odometry in
  low light using binary descriptors,'' {\em IEEE Robotics and Automation
  Letters}, vol.~2, no.~2, pp.~444--451, 2017.

\bibitem{aerconf2018}
{F. Mascarich, S. Khattak, C. Papachristos, K. Alexis}, ``A multi-modal mapping
  unit for autonomous exploration and mapping of underground tunnels,'' in {\em
  IEEE Aerospace Conference}, March 2018.

\bibitem{BRAND}
E.~R. Nascimento {\em et~al.}, ``Brand: A robust appearance and depth
  descriptor for rgb-d images,'' in {\em International Conference on
  Intelligent Robots and Systems}, 2012.

\bibitem{RISAS}
K.~Wu {\em et~al.}, ``Risas: A novel rotation, illumination, scale invariant
  appearance and shape feature,'' in {\em International Conference on Robotics
  and Automation}, 2017.

\bibitem{levinson2013}
J.~Levinson and S.~Thrun, ``Automatic online calibration of cameras and
  lasers.,'' in {\em Robotics: Science and Systems}, vol.~2, 2013.

\bibitem{ORB}
E.~Rublee, V.~Rabaud, K.~Konolige, and G.~Bradski, ``Orb: An efficient
  alternative to sift or surf,'' in {\em International conference on Computer
  Vision (ICCV)}, 2011.

\bibitem{realsense}
L.~Keselman {\em et~al.}, ``Intel (r) realsense (tm) stereoscopic depth
  cameras,'' in {\em Computer Vision and Pattern Recognition Workshops
  (CVPRW)}, IEEE, 2017.

\bibitem{kalibr}
P.~Furgale, J.~Maye, J.~Rehder, T.~Schneider, and L.~Oth, ``Kalibr.''
  \url{https://github.com/ethz-asl/kalibr}, 2014.

\bibitem{mpc_rosbookchapter}
{M. Kamel, T. Stastny, K. Alexis, and R. Siegwart}, ``Model predictive control
  for trajectory tracking of unmanned aerial vehicles using ros,'' {\em
  Springer Book on Robot Operating System (ROS)}.

\end{thebibliography}
